\documentclass[letterpaper, 10 pt, conference]{ieeeconf}  % Comment this line out if you need a4paper

\IEEEoverridecommandlockouts                              % This command is only needed if 
\usepackage{amsmath,amssymb,amsfonts}
\usepackage{algorithmic}
\usepackage{graphicx}
\usepackage{textcomp}
\usepackage{url}
\usepackage{xcolor}
\usepackage{calc}
\usepackage{verbatim}
\usepackage{multirow}
\usepackage{booktabs}   % \toprule, \midrule, \bottomrule, \cmidrule
\usepackage{multirow}   % if using \multirow
\usepackage{comment}    % because you use \begin{comment}
\usepackage{amsmath,amssymb,amsfonts}
\usepackage{algorithmic}
\usepackage{graphicx}
\usepackage{textcomp}
\usepackage{xcolor}
\usepackage{verbatim}
\title{\LARGE \bf
ClothCode: Generative Modeling for Garment State Estimation and Dynamics Prediction
}

\author{
Yingdong Ru,
Lipeng Zhuang,
Shiyu Fan,
Zhuo He,\\
Florent P. Audonnet,
Gerardo Aragon Camarasa%
\thanks{The authors are with the University of Glasgow, Glasgow, United Kingdom.
\texttt{2649534r@student.gla.ac.uk}}
}

\begin{document}

\maketitle
\thispagestyle{empty}
\pagestyle{empty}

%%%%%%%%%%%%%%%%%%%%%%%%%%%%%%%%%%%%%%%%%%%%%%%%%%%%%%%%%%%%%%%%%%%%%%%%%%%%%%%%
\begin{abstract}
We introduce \textbf{ClothCode}, a unified generative framework for
garment state estimation and action-conditioned dynamics prediction in a
shared discrete configuration space. A material-aligned tokenizer encodes
garment configuration as residual vector-quantized patch tokens, while
task-specific Masked Generative Transformers predict these tokens through
parallel confidence-guided refinement. Across three garment templates,
ClothCode reduces mean displacement error by 26.6\% for state estimation
and 37.0\% for dynamics prediction relative to the strongest baselines.
Dynamics prediction requires only 16.7\,ms per window, compared with
556.2\,ms for diffusion. Integrated into the same model-predictive
controller, ClothCode improves real-world folding and flattening success
from 56.7\% to 78.9\%. Further results and videos are available at: \url{https://anonymous.4open.science/r/ClothCode-DE32}
\end{abstract}

%%%%%%%%%%%%%%%%%%%%%%%%%%%%%%%%%%%%%%%%%%%%%%%%%%%%%%%%%%%%%%%%%%%%%%%%%%%%%%%%
\begin{figure*}[t]
\vspace{1mm}
    \centering
    \includegraphics[width=0.85\linewidth]{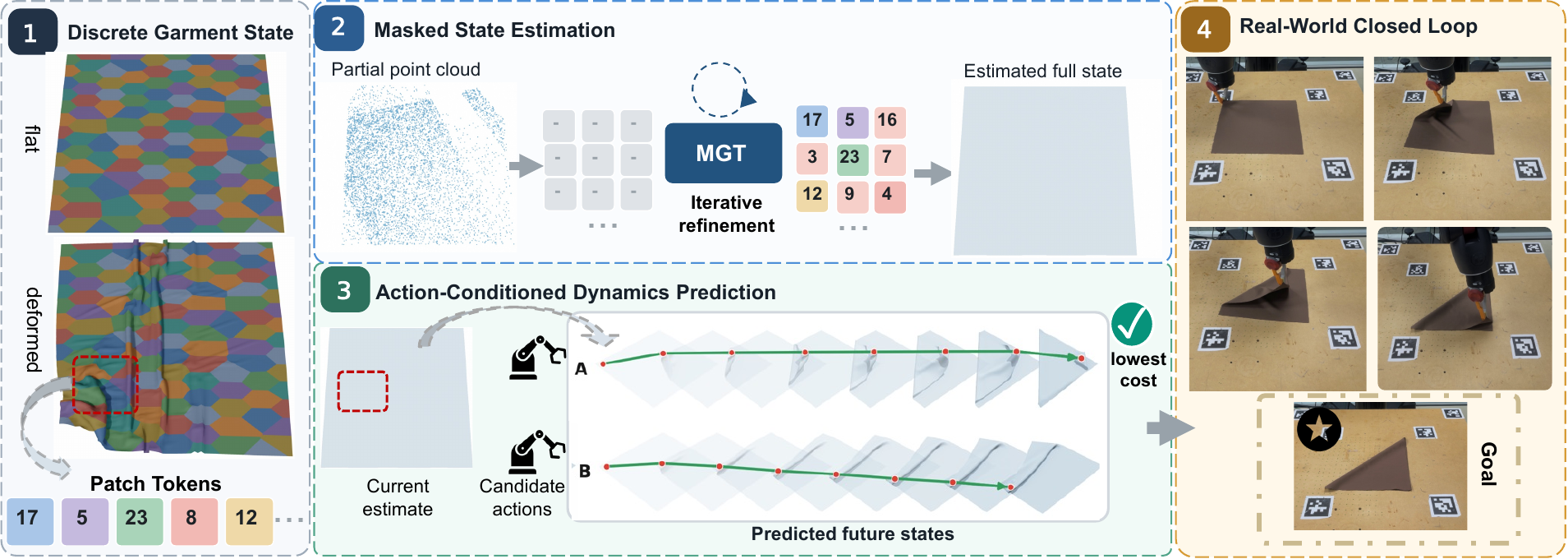}
    
    \vspace{-6pt}
    \caption{\textbf{Overview of our discrete generative framework for model-based garment manipulation.}
A garment configuration is encoded as a set of discrete patch tokens shared by two Masked Generative Transformers (MGTs).
\textbf{(1)} The discrete garment representation captures local deformation states over fixed material patches.
\textbf{(2)} For state estimation, an MGT iteratively predicts the complete token configuration from a partial point cloud.
\textbf{(3)} For dynamics prediction, an action-conditioned MGT predicts future garment configurations for candidate actions, allowing MPC to select the trajectory with the lowest goal cost.
\textbf{(4)} The resulting perception--prediction loop enables closed-loop garment manipulation on a real robot.}
    \label{fig:main1}
    \vspace{-5mm}
\end{figure*}

\section{Introduction}
Model-based cloth manipulation involves robots inferring the full shape of a garment from partial views and forecasting how it will deform in response to their actions~\cite{longhini2025unfolding}. These tasks are challenging because garments have high-dimensional configuration spaces and undergo complex deformations with frequent self-contact and enviromental interactions. Severe self-occlusion makes the current state ambiguous, while unknown material properties and contact conditions add uncertainty to the garment's future behavior. Effective planning thus requires an accurate garment-state estimation and an efficient dynamics model that can be queried repeatedly to evaluate candidate actions.

% Existing deterministic cloth models produce a single reconstruction or future state~\cite{chi2021garmentnets,huang2022medor,zhang2024adaptigraph}, limiting their ability to capture the multiple configurations or future outcomes that are consistent with the same inputs.  
% Existing deterministic cloth models map each input to a single predicted state~\cite{chi2021garmentnets,huang2022medor,zhang2024adaptigraph}, without explicitly modeling the conditional distribution over garment configurations. 

Existing cloth state and dynamics models are predominantly deterministic, producing a single prediction for each input~\cite{chi2021garmentnets,huang2022mesh,zhang2024adaptigraph}. Under ambiguous targets, deterministic predictors optimized with point-wise losses can favor conditional averages rather than distinct plausible garment configurations. UniClothDiff~\cite{tian2025diffusion} instead formulates both state estimation and dynamics prediction as conditional diffusion. However, it generates dense continuous vertex states and displacements without explicitly constraining the predictions to learned local deformation representations. Its sampler also repeatedly denoises the complete output, which makes repeated evaluation costly during planning.

To address these limitations, we introduce \textbf{ClothCode}, a unified generative framework that formulates garment-state estimation and dynamics prediction as conditional generation in a shared discrete garment space as shown in Fig.~\ref{fig:main1}. Our working hypothesis is that, although complete garment configurations vary widely, local deformation patterns are shared across different material regions and garment states. We thus partition the mesh into fixed, material-aligned patches and encode each patch as discrete tokens using residual vector quantization~\cite{van2017neural,lee2022autoregressive}, with codebooks shared across patches. A topology-aware decoder maps these tokens back to a continuous mesh geometry, resulting into a compact prediction space of learned patch-level configurations. For this, we use task-specific Masked Generative Transformers (MGTs) for state estimation and dynamics prediction~\cite{chang2022maskgit,guo2024momask,zhuang2026masked}, which iteratively refine tokens using confidence scores to enable parallel generation.

In this paper, our central contribution is to \textit{show that a shared, material-aligned discrete representation, combined with masked iterative generation, can make both garment perception and dynamics prediction more accurate, substantially faster, and more effective for real-world, closed-loop robotic manipulation}. We demonstrate this on a large simulated dataset of 134,400 trajectories across three garment types, and validate it through real-world folding and flattening experiments on a dual-arm robot. Specifically, we found that ClothCode reduces mean displacement error by 26.6\% (state) and 37.0\% (dynamics) versus baselines, cuts dynamics inference time from 556.2,ms to 16.7,ms per window in simulation environment, and boosts real-world manipulation success from 56.7\% to 78.9\%.

\section{Related Work}

\noindent\textbf{Cloth State Estimation}
Recovering complete cloth geometry from partial observations has been studied through canonical-space completion, mesh reconstruction, and generative modeling. GarmentNets~\cite{chi2021garmentnets} performs category-level garment reconstruction through canonical-space completion, while MEDOR~\cite{huang2022mesh} further refines reconstructed meshes using test-time self-supervision. More recently, UniClothDiff~\cite{tian2025diffusion} formulates full-state estimation as conditional diffusion. These methods represent the recovered garment state in continuous geometric or latent spaces. In contrast, our discrete bottleneck presented herein constrains prediction to configurations expressed via a learned code vocabulary, while categorical token predictions further enable confidence-guided masked completion.

\noindent\textbf{Cloth Dynamics Prediction}
Learned cloth dynamics have been modeled using particle-, graph-, and generative representations\cite{pfaff2020learning,shi2024robocraft,jiang2025phystwin,lin2022learning}. Particle-based graph neural dynamics predict object motion through relational message passing~\cite{zhang2024adaptigraph}. PGND~\cite{zhang2025particle} instead combines object particles with a spatial grid to capture both dense local motion and global spatial structure UniClothDiff~\cite{tian2025diffusion} models garment dynamics using conditional diffusion over continuous future states, while ClothTransformer~\cite{zhang2026clothtransformer} performs autoregressive prediction in a learned continuous latent space. In contrast, our model jointly refines discrete patch--time configuration tokens within each prediction window, allowing low-confidence future components to be revised without state-by-state autoregressive generation.

\noindent\textbf{Discrete and Masked Generative Modeling}
Vector quantization provides discrete latent representations, with residual vector quantization increasing capacity through successive residual coding~\cite{van2017neural}~\cite{lee2022autoregressive}\cite{esser2021taming}. MaskGIT~\cite{chang2022maskgit} introduced parallel masked generation with confidence-guided token refinement\cite{chang2023muse,yu2023magvit}, which has since been extended to structured sequential domains such as 3D human motion with MoMask~\cite{guo2024momask} and robot action generation with MGP~\cite{zhuang2026masked}. Discrete abstractions have also been explored for cloth, but primarily as manipulation-oriented topological or semantic states~\cite{tzelepis2022semantic} rather than representations of complete fine-grained geometry. In this paper, we propose to learn patch-wise discrete tokens for the full non-rigid patch-wise garment configuration and use the same configuration space for both full-state estimation and action-conditioned dynamics prediction.

% \subsection{Garment Manipulation Datasets}

% Existing garment datasets lack several forms of supervision required to jointly study perception and dynamics. GarmentNets~\cite{chi2021garmentnets} provides complete geometry for reconstruction but not dense action-conditioned manipulation trajectories or controlled physical variation. Flat'n'Fold provides real manipulation demonstrations but lacks temporally dense complete-mesh ground truth with persistent material correspondence~\cite{zhuang2025flatnfold}. RGBench~\cite{hu2026rgbench} targets simulator fidelity and provides only a small real-world evaluation set rather than large-scale training trajectories, while ClothTransformer~\cite{zhang2026clothtransformer} offers topology-diverse simulations but few repeated manipulations per topology and no paired partial observations or systematic physics variation. UniClothDiff~\cite{tian2025diffusion} considers both state estimation and dynamics, but its training data and generation pipeline are not publicly released. Our dataset combines partial observations, dense correspondence-preserving meshes, robot actions, repeated manipulation trajectories, and controlled physical variations, together with separate action- and physics-OOD splits. This enables unified evaluation of full-state estimation, dynamics and world-model learning, physical adaptation, and model-based control.

\vspace{-3mm}
\section{Method}

\label{sec:method}

\begin{figure*}[t]
\vspace{1mm}
    \centering
    \includegraphics[width=0.92\linewidth]{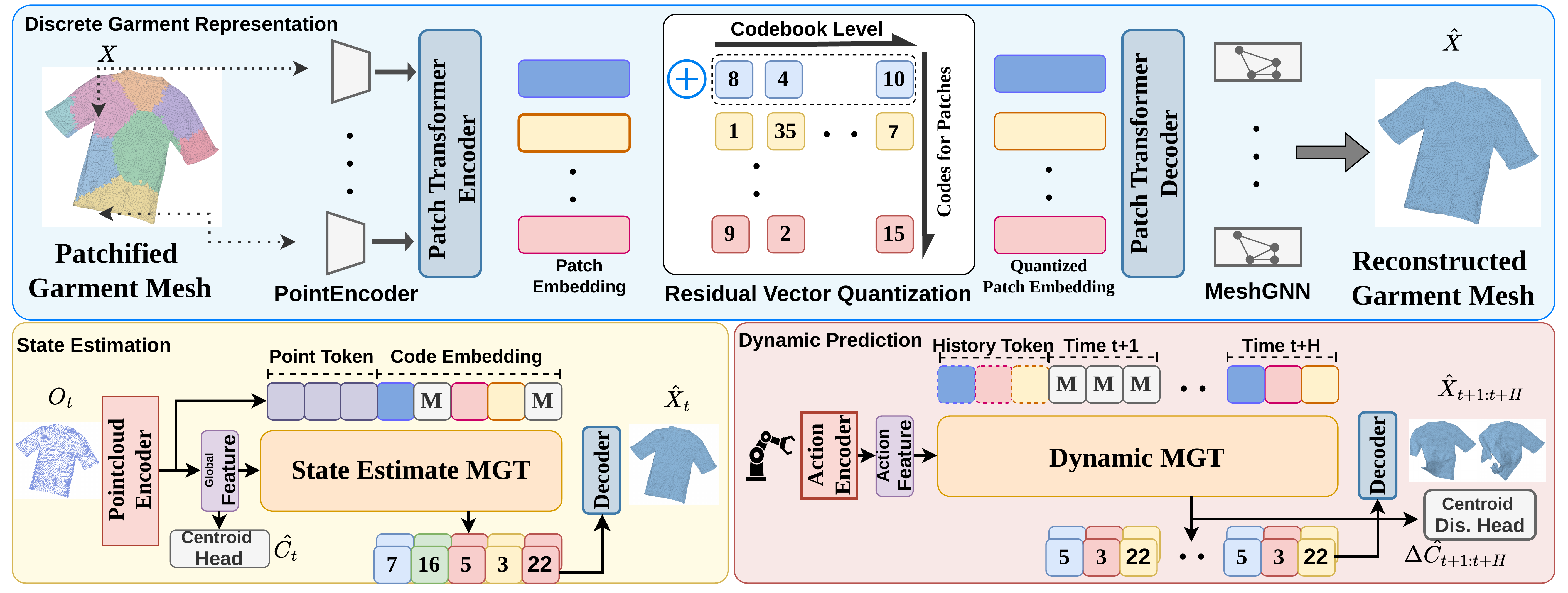}
    %\vspace{-6pt}
   \caption{\textbf{Overview of ClothCode.} ClothCode first learns a discrete representation of garment shape (top). Each mesh is divided into fixed patches that track the same material regions across deformations, encoded as two-level residual vector-quantization (RVQ) tokens, and reconstructed by a shared mesh decoder. This frozen representation provides a common prediction space for two tasks. For state estimation (bottom left), a masked transformer combines a partial point cloud with masked garment tokens to recover the complete garment configuration and its global position. For dynamics prediction (bottom right), a second masked transformer combines recent garment states with a candidate end-effector trajectory and jointly predicts the future garment sequence and its global motion. $\mathrm{M}$ denotes a masked token.}

    \label{fig:main2}
    \vspace{-5mm}
\end{figure*}

Given a known garment template with fixed connectivity and material correspondence, we represent its configuration at time \(t\) by a triangular mesh with \(V\) vertices,
$\mathbf{X}_t=[\mathbf{x}_{t,1},\ldots,\mathbf{x}_{t,V}]^\top
\in\mathbb{R}^{V\times3}$, where $\mathbf{x}_{t,v}$ denotes the position
of vertex $v$. Because vertex indexing is preserved across configurations,
each vertex corresponds to the same material point over time. As stated above, ClothCode addresses two tasks: \emph{garment state estimation}, which
reconstructs the complete garment configuration from a partial point cloud, and \emph{dynamics prediction}, which predicts future
configurations from recent configurations together with the planned
end-effector motion.

% During training and for each garment template, we first learn
% a garment tokenizer on complete simulated meshes that encodes mesh geometry
% relative to its centroid as discrete codebook indices, which we term
% \emph{garment tokens} (Sec.~\ref{sec:representation}). The tokenizer is then
% frozen, defining the discrete configuration space shared by two conditional
% Masked Generative Transformers (MGTs), one for state estimation and one for
% dynamics prediction (Sec.~\ref{sec:generative_models}). At inference, each
% MGT predicts the masked tokens in parallel and then iteratively remasks and
% re-predicts the least confident tokens (Sec.~\ref{sec:inference}); the frozen tokenizer
% decoder maps the predicted tokens back to centroid-relative mesh geometry.
%All models are trained separately for each garment template.
As shown in Fig.~\ref{fig:main2}, for each template, we first train a garment tokenizer on complete
simulated meshes to encode centroid-relative garment configuration as discrete codebook
indices, which we term \emph{garment tokens} (Sec.~\ref{sec:representation}).
The tokenizer is then frozen, defining a shared discrete configuration space
for two conditional Masked Generative Transformers (MGTs) for state estimation and dynamics prediction.
%Because the token space represents garment shape relative to its centroid, global translation is recovered separately by lightweight task-specific centroid heads. 
Since each mesh is centered before tokenization, the garment tokens encode only centroid-relative geometry. Global translation is therefore predicted separately by task-specific translation heads (Sec.~\ref{sec:generative_models}).
At inference, both MGTs predict masked tokens in parallel and iteratively re-mask and re-predict low-confidence tokens (Sec.~\ref{sec:inference}); the frozen decoder maps the final tokens back to centroid-relative mesh configurations, which are combined with the corresponding centroid estimates to recover the garment configurations in world coordinates.

\subsection{Discrete Garment Configuration}
\label{sec:representation}
%Our tokenizer combines a patch encoder, a mesh decoder, and residual vector quantization (RVQ). The encoder summarizes deformation within and across material regions, RVQ converts these features into discrete codes, and the decoder integrates the quantized features to reconstruct the complete mesh.

\paragraph{Patch construction}
For tokenization, we remove global translation and normalize each mesh by
defining $\mathbf{c}_t=\frac{1}{V}\sum_{v=1}^{V}\mathbf{x}_{t,v}$ and
$\tilde{\mathbf{x}}_{t,v}=(\mathbf{x}_{t,v}-\mathbf{c}_t)/s_x$, where
$s_x$ is a fixed normalization scale which is computed once from the training set as the standard deviation of centroid-centered vertex coordinates. The resulting normalized
centroid-relative mesh is denoted by $\tilde{\mathbf{X}}_t$.
%We then partition each garment template once in its flat reference configuration and keep patch membership fixed across deformation.\lp{need revise} 
We partition the flat reference mesh into \(P\) patches and keep the same vertices in each patch across all garment configurations.
%We use farthest-point sampling to select $P$ patch seeds on the reference mesh and assign each vertex to its nearest seed, yielding $P$ patches indexed by $p=1,\ldots,P$.
Specifically, farthest-point sampling selects $P$ representative vertices on the reference mesh as patch centers, and each vertex is assigned to its nearest center, yielding $P$ patches indexed by $p=1,\ldots,P$.
Because vertex correspondence is preserved over time, each patch therefore represents the same material region across all garment configurations.

% \paragraph{Patch construction.}
% To obtain a compact token sequence while preserving material correspondence, we partition each garment template once in its flat reference configuration and keep patch membership fixed across deformation. For each garment template, we use its flat undeformed mesh $\mathbf{X}^{0}$ as a fixed reference mesh, while $\mathbf{X}_t$ denotes the garment configuration to be encoded. Both meshes are centered at their respective vertex centroids and divided by the same fixed isotropic scale $s_x$. We denote the resulting normalized meshes by $\widetilde{\mathbf{X}}^{0}$ and $\widetilde{\mathbf{X}}_t$. Farthest-point sampling on the reference mesh selects $P$ representative vertices as partition seeds, and each vertex is assigned to its nearest seed, defining the fixed vertex set $\mathcal{V}_p$ of patch $p$. Because this partition is computed once on the reference mesh and vertex correspondence is preserved, each patch contains the same material vertices across all garment configurations. We define the reference center of each patch as $\boldsymbol{\mu}^{0}_{p}=|\mathcal{V}_p|^{-1}\sum_{v\in\mathcal{V}_p}\widetilde{\mathbf{x}}^{0}_{v}$, and $\mathcal{N}(p)$ as the six nearest patches according to the
% Euclidean distance between their reference centers.

\paragraph{Patch encoder}
For each patch, we encode both its motion relative to the flat reference configuration and its within-patch deformation. The patch-level feature contains the displacement of the patch center, the displacements of its neighboring patches, and its center position in the reference mesh. In parallel, a shared PointNet-style encoder~\cite{qi2017pointnet} aggregates the local deformation of the vertices within the patch. The two features are concatenated and projected to the model dimension. After adding a learned patch embedding, the \(P\) patch features are processed together by a Transformer using self-attention~\cite{vaswani2017attention}, followed by a linear projection to obtain a continuous patch latent, \(\mathbf{z}_{t,p}\in\mathbb{R}^{d_z}\).

\paragraph{Residual Vector Quantization (RVQ)}
Each continuous patch latent $\mathbf{z}_{t,p}$ is discretized by a
two-level residual vector quantizer (RVQ)
~\cite{van2017neural,guo2024momask}, yielding two discrete codebook indices
per patch. Each level $r\in\{1,2\}$ uses a separate codebook,
$\{\mathbf{e}^{(r)}_k\}_{k=1}^{K}$, shared across all patches of a garment
template. 
Starting from
$\mathbf{r}^{(1)}_{t,p}=\mathbf{z}_{t,p}$, each level thereafter selects the nearest
codebook vector for the current residual. That is,
% \begin{equation}
% q^{(r)}_{t,p}
% =\arg\min_{k\in\{1,\ldots,K\}}
% \left\|\mathbf r^{(r)}_{t,p}-\mathbf e^{(r)}_k\right\|_2^2,
% \end{equation}
% \begin{equation}
% \mathbf r^{(r+1)}_{t,p}
% =\mathbf r^{(r)}_{t,p}
% -\mathbf e^{(r)}_{q^{(r)}_{t,p}}.
% \label{eq:rvq}
% \end{equation}
at each stage $r$, the residual $\mathbf r^{(r)}_{t,p}$ is approximated by the nearest codeword $\mathbf e^{(r)}_{q^{(r)}_{t,p}}$ from a learned codebook, where $q^{(r)}_{t,p}$ minimizes the squared Euclidean distance. This codeword is subtracted to form the next residual $\mathbf r^{(r+1)}_{t,p}$, and the process repeats across stages, progressively representing the original vector as a sum of discrete codewords. The selected codebook vectors are summed to form the quantized patch latent
$\mathbf{z}^{q}_{t,p}
=\sum_{r=1}^{2}\mathbf{e}^{(r)}_{q^{(r)}_{t,p}}$, which is passed to the
mesh decoder. We refer to the two indices
$\{q^{(1)}_{t,p},q^{(2)}_{t,p}\}$ as the garment tokens of patch $p$.
Collecting them across all patches gives the discrete garment configuration
$\mathbf{Q}_t\in\{1,\ldots,K\}^{P\times 2}$.

%\paragraph{Mesh decoder.}
%The decoder maps the quantized patch latents back to the normalized garment mesh. A Transformer encoder first contextualizes the $P$ patch features.
%Each patch feature is then broadcast to its assigned vertices and concatenated with a Fourier encoding\cite{tancik2020fourier} of each vertex's offset from the patch center in the flat reference mesh. A mesh GNN propagates these features along the fixed material connectivity and predicts a three-dimensional displacement for each vertex relative to the reference mesh, yielding the reconstructed configuration $\widehat{\widetilde{\mathbf{X}}}_t$.

\paragraph{Mesh decoder}
The decoder reconstructs the normalized garment mesh from the quantized patch latents. A Transformer first jointly processes the \(P\) quantized patch features to exchange information across material patches before vertex-level reconstruction. For each vertex, we concatenate its corresponding patch feature with a Fourier positional encoding\cite{tancik2020fourier} of its offset from the patch center in the flat reference mesh.
A mesh GNN propagates the vertex features along the fixed mesh connectivity and predicts a 3-dimensional displacement for each vertex relative to the reference mesh, yielding the reconstructed configuration $\widehat{\widetilde{\mathbf{X}}}_t$.

% \paragraph{Mesh decoder.}
% The decoder reconstructs the normalized mesh from the quantized patch
% features. A bidirectional Transformer first exchanges information across the $P$
% quantized patch features, allowing each patch to be reconstructed in the
% context of the complete garment configuration. 
% \fan{The default transformer is bidirectional and time-related, no need to mention too much.}
% The resulting feature of each patch is then broadcast to
% its assigned vertices and concatenated with a Fourier encoding of the
% reference local coordinate $\boldsymbol{\ell}^{0}_v$, providing
% vertex-specific information within each patch. \fan{SF: Same problems. Too fussy. Just say the vertices features then are sent into the MeshGNN with fourier positional embedding.} A mesh GNN subsequently
% performs message passing along the fixed material connectivity, avoiding
% connections between spatially adjacent but materially disconnected folded
% layers. For each mesh edge, we use its reference edge vector and length as
% edge features, since the deformed vertex positions are the prediction
% target. The decoder finally predicts a three-dimensional displacement for
% each vertex relative to its reference position, yielding the reconstructed
% normalized mesh $\widehat{\widetilde{\mathbf{X}}}_t$.

\paragraph{Training}
The tokenizer, comprising the patch encoder, RVQ module, and mesh decoder, is trained to reconstruct the normalized input mesh. The training objective combines mesh reconstruction and RVQ commitment losses, i.e. $\mathcal{L}_{\mathrm{tok}}=\mathcal{L}_{\mathrm{rec}}+\beta\mathcal{L}_{\mathrm{commit}}$, 
where $\mathcal{L}_{\mathrm{rec}}$ is the mean-squared error between
$\widehat{\widetilde{\mathbf{X}}}_t$ and
$\widetilde{\mathbf{X}}_t$, and $\beta$ weights the commitment loss.
The RVQ codebooks are updated using exponential moving averages.
After training, the patch encoder, RVQ codebooks, and mesh decoder are frozen.

\subsection{Masked Generative Garment Models}
\label{sec:generative_models}

We formulate state estimation and dynamics prediction as conditional masked generation in the discrete garment space, i.e. 
$p(\mathbf Q_t\mid\mathbf O_t)$ and $p(\mathbf Q_{t+1:t+H}\mid\mathbf Q_{t-C+1:t},\mathbf A)$, respectively, where $\mathbf{O}_t$ denotes the partial observation and $\mathbf{A}$ the planned end-effector motion. 
Both tasks are modeled by task-specific Masked Generative Transformers (MGTs)~\cite{chang2022maskgit,zhuang2026masked}, each comprising a conditioning encoder and a masked Transformer operating in the discrete garment space. 
%Given the task condition and a masked garment configuration, the codebook embedding corresponding to each masked RVQ index is replaced by a learned level-specific $[\mathrm{MASK}]$ embedding. 
%At each refinement iteration, the MGT predicts the masked indices at both RVQ levels in parallel; high-confidence predictions are retained, while low-confidence indices are re-masked for the next iteration. 
%The final predicted indices form the discrete garment configuration $\mathbf Q$, which is dequantized using the frozen RVQ codebooks and decoded by the frozen mesh decoder to reconstrcut the garment configuration.
Starting from masked target garment tokens, the MGT predicts their codebook indices in parallel and iteratively re-predicts low-confidence tokens over a small number of refinement steps. The predicted indices are then converted to quantized patch latents using the frozen RVQ codebooks and decoded by the mesh decoder to reconstruct the garment configuration.

%Both MGTs predict the two RVQ indices of each target patch through separate output heads.

% Their code embeddings are combined into a single patch representation, while masked code variables are replaced by learned level-specific $[\mathrm{MASK}]$ embeddings. The Transformer jointly models the resulting patch representations, and level-specific prediction heads output categorical distributions over the corresponding codebooks. Masking and prediction operate on individual RVQ code variables, allowing uncertain codes at any residual level to be iteratively re-predicted while retaining confident predictions at the others. The resulting multi-level code indices form the discrete garment configuration $\mathbf{Q}$, which is decoded into mesh geometry by the frozen tokenizer decoder.

%\lp{need revise later}
\paragraph{State estimation}
\label{sec:stateestimation}
Given a partial point-cloud observation $\mathbf{O}_t$, let
$\mathbf{c}^{\mathrm{vis}}_t$ denote the centroid of the observed points. We center the point cloud at $\mathbf{c}^{\mathrm{vis}}_t$ and normalize its coordinates using the same fixed scale $s_x$ as the garment tokenizer. The point-cloud encoder produces two outputs with different roles. It maps the observed points to point features. A fixed set of learned query tokens attends to these point features through
cross-attention to produce multiple point-cloud tokens. In parallel, the point features are also pooled into a single global point-cloud feature that captures the overall observation. For each garment patch, the embeddings of its two RVQ levels are summed to form one patch token. The point-cloud tokens are prepended to the garment patch tokens, and the combined sequence is processed by a stack of self-attention Transformer blocks. The global point-cloud feature modulates each Transformer block through AdaLN-zero~\cite{peebles2023scalable}. The output feature of each garment patch is passed to two level-specific classification heads to predict its two RVQ indices.
The global point-cloud feature is also used for centroid prediction.
Together with $\mathbf{c}^{\mathrm{vis}}_t$, it is passed to a centroid MLP head
that predicts the centroid residual $\widehat{\boldsymbol{\delta}}_t$.
The complete garment centroid is then recovered as
$\widehat{\mathbf{c}}_t=
\mathbf{c}^{\mathrm{vis}}_t+s_x\widehat{\boldsymbol{\delta}}_t$.

%\paragraph{State estimation.}
%Given a partial point-cloud observation $\mathbf{O}_t$, let
%$\mathbf{c}^{\mathrm{vis}}_t$ denote the centroid of the observed points.
%We center the point cloud at $\mathbf{c}^{\mathrm{vis}}_t$ and normalize
%its coordinates using the same fixed scale $s_x$ as the garment tokenizer.

%The point-cloud encoder produces two outputs with different roles. It first
%maps the observed points to point features. A fixed set of learned queries
%uses cross-attention over these point features to produce multiple point-cloud
%tokens, which retain information about the observed geometry. The point
%features are also pooled into a single global point-cloud feature that
%summarizes the observation.

%For each garment patch, the embeddings of its two RVQ levels are summed to
%form one patch token. The point-cloud tokens are prepended to the garment
%patch tokens, and the combined sequence is processed by a stack of
%self-attention Transformer blocks. The global point-cloud feature additionally
%The output feature of each garment patch is passed to two level-specific
%classification heads to predict its two RVQ indices.

%The global point-cloud feature is also used for centroid prediction.
%Together with $\mathbf{c}^{\mathrm{vis}}_t$, it is passed to a centroid head
%that predicts the centroid residual $\widehat{\boldsymbol{\delta}}_t$,
%from which the complete garment centroid is recovered as
%$\widehat{\mathbf{c}}_t=
%\mathbf{c}^{\mathrm{vis}}_t+s_x\widehat{\boldsymbol{\delta}}_t$.

\paragraph{Dynamics prediction.}
\label{sec:dynamicprediction}
The dynamics model predicts \(H\) future discrete garment configurations conditioned on \(C\) context configurations and the planned end-effector trajectory. At each future step, the action input combines the gripper position relative to the garment centroid at the final context frame with the end-effector displacement between the current and previous trajectory steps. Both are normalized and encoded using Fourier features~\cite{tancik2020fourier} followed by an MLP, and the resulting embedding is added to the garment patch tokens at the corresponding future step.  
Together, the (C) context steps and (H) action-conditioned future steps are concatenated along the temporal dimension to form a joint spatiotemporal sequence. Each of the \(C+H\) time steps contains \(P\) garment patch tokens, yielding \((C+H)P\) patch--time positions.
%The dynamics model predicts $H$ future discrete garment configurations
%conditioned on $C$ context configurations and the planned end-effector motion. At each future step, the action condition comprises the gripper position relative to the garment centroid from the final context frame and the prescribed end-effector displacement for that step. After normalization, these quantities are encoded using Fourier features~\cite{tancik2020fourier} followed by an MLP. The resulting action embedding is added to the garment patch representations at the corresponding future step.
%For each of the $C+H$ time steps, the model contains $P$ garment patch tokens, giving $(C+H)P$ patch--time positions. 
Each token is assigned a learned patch embedding that identifies the garment patch and a learned temporal embedding that identifies the time step. A separate role embedding indicates whether the token belongs to a context or future configuration. The resulting $(C+H)P$ garment patch tokens are flattened into a single sequence and processed by a stack of self-attention Transformer blocks in MGT, allowing each token to attend to garment patches across all context and future time steps. Two level-specific classification heads predict the two RVQ indices from the Transformer output at each future patch--time position.  Since the discrete garment configuration represents centroid-relative geometry, global translation is predicted separately. 
%The hidden states at the future positions are pooled across garment patches at each time step and passed to a centroid-displacement head that predicts the cumulative centroid displacement from the final context configuration.
The centroid-displacement head uses the Transformer outputs obtained with all future garment tokens masked. For each future time step, the outputs of its $P$ garment patches are averaged and passed to the head to predict the centroid displacement relative to the final context configuration.

\paragraph{Training }

%Both MGTs are trained by masked-token prediction. For each example, we sample $u\sim\mathcal{U}(0,1)$ and independently mask target tokens with probability $\gamma(u)=\cos(\pi u/2)$. 
%The centroid regression heads are trained jointly with the corresponding masked-token objectives:

During training, for each example, we sample
$u\sim\mathcal{U}(0,1)$ and set the mask ratio to
$\gamma(u)=\cos(\pi u/2)$. We randomly mask this fraction of the target RVQ
indices, and each masked index is replaced by the learned $[\mathrm{MASK}]$
embedding of its RVQ level. Let $\mathcal{M}$ denote the masked set and $\boldsymbol{\eta}$ the task
condition. For state estimation, $\boldsymbol{\eta}=\mathbf{O}_t$; for
dynamics prediction,
$\boldsymbol{\eta}=(\mathbf{Q}_{t-C+1:t},\mathbf{A})$, with all context tokens kept unmasked. Given the task condition and the remaining unmasked tokens, ClothCode predicts the original indices at the masked positions. We minimize cross-entropy over the masked tokens:
\begin{equation}
\mathcal{L}_{\mathrm{mask}}
=
-\frac{1}{|\mathcal{M}|}
\sum_{i\in\mathcal{M}}
\log p_{\theta}
\left(
q_i \mid q_{\bar{\mathcal{M}}},\boldsymbol{\eta}
\right),
\label{eq:masked_loss}
\end{equation}
where $q_{\bar{\mathcal{M}}}$ denotes the unmasked target tokens. The masked-token objective is optimized jointly with an $L_1$ loss for centroid prediction:
\begin{equation}
\begin{aligned}
\mathcal{L}_{\mathrm{state}}
&=
\mathcal{L}_{\mathrm{mask}}^{\mathrm{state}}
+\frac{\lambda_c}{3}
\left\|
\widehat{\boldsymbol{\delta}}_t-\boldsymbol{\delta}_t
\right\|_1,\\
\mathcal{L}_{\mathrm{dyn}}
&=
\mathcal{L}_{\mathrm{mask}}^{\mathrm{dyn}}
+\frac{\lambda_{\mathrm{trans}}}{3H}
\sum_{h=1}^{H}
\left\|
\widehat{\mathbf{d}}_{t+h}-\mathbf{d}_{t+h}
\right\|_1 .
\end{aligned}
\label{eq:task_losses}
\end{equation}
where, $\lambda_c$ and $\lambda_{\mathrm{trans}}$ are the weights for the centroid-residual and future-translation $L_1$ losses, respectively. The factor $1/3$ averages the absolute error over the three spatial coordinates, while $1/(3H)$ additionally averages over the $H$ future steps. For dynamics prediction, $\widehat{\mathbf d}_{t+h}$ is obtained from the fully masked companion pass described above.

\vspace{-2.5mm}
\subsection{Mask-and-Refine Inference}
\label{sec:inference}

Both tasks use the same confidence-guided mask-and-refine procedure from MGT\cite{chang2022maskgit}. State estimation begins with all garment tokens masked, whereas dynamics prediction masks all future tokens while keeping the context tokens fixed. We use four refinement iterations for both tasks. At each iteration, MGT predicts the logits of all masked tokens in parallel, from which the codebook indices are sampled using the Gumbel-Max with temperature $\tau$. The softmax probability assigned to each sampled codebook index is used as its confidence score. A decreasing number of the lowest-confidence tokens are re-masked according to a cosine schedule and generated again in the next iteration, while the higher-confidence tokens are retained. After refinement, the predicted indices are converted to quantized patch latents using the frozen RVQ codebooks and decoded by the frozen mesh decoder. The meshes are then transformed to world coordinates using the estimated garment centroid for state estimation and the predicted centroid displacements for dynamics prediction.

\vspace{-1.5mm}
\subsection{Implementation Details.}
We partition each garment template into $P=128$ material-aligned patches and use a two-level RVQ with $K=512$ entries per level and a $64$-D patch latent. The tokenizer uses a $256$-D Transformer backbone with two encoder and four decoder layers. The dynamics model uses four Transformer blocks with eight attention heads and a hidden dimension of $256$, and predicts $H=10$ future configurations from $C=3$ context configurations. The state-estimation model uses six Transformer blocks with eight attention heads and the same hidden dimension. We use four mask-and-refine iterations for dynamics prediction and state estimation.

\section{Experiment Setup}
\subsection{Dataset Collection.}

Reliable cloth state estimation and dynamics prediction require data spanning diverse cloth configurations, manipulation actions, and physical properties. We therefore construct a large-scale simulated dataset in Isaac Sim 5.1 with Isaac Lab 2.2.1~\cite{mittal2025isaac}, using GPU-based deformable-body simulation with self-collision enabled. We consider three fixed-topology templates---a fabric sheet, a T-shirt, and a sweater---initialized from flat, crumpled, and folded configurations.

Each action instance is parameterized by a vector controlling the grasp, target, lifting height, and motion, and records the full garment mesh and end-effector Cartesian position at every frame. We include \emph{fold}, \emph{reposition}, and \emph{flatten} actions, together with garment-specific \emph{sleeve-fold} and \emph{cross-arm} actions. For each valid template--initial-configuration--action combination, we sample 700 distinct action instances. Each instance is simulated under eight material-property configurations spanning friction, stiffness, and density, with additional perturbations within each configuration to broaden the range of plausible cloth properties. The resulting dataset contains 16,800 action instances and 134,400 simulated trajectories. For state estimation, partial point clouds are rendered from virtual depth cameras. We randomize camera azimuth, elevation, and distance, and include the gripper and forearm in the rendered scene to model robot-induced occlusion. We further apply zero-mean Gaussian depth noise with \(\sigma=3\) mm, randomly drop \(10\%\)--\(20\%\) of observed points, and perturb camera extrinsics with per-axis Gaussian translation noise of \(\sigma=5\) mm and random-axis rotation noise of \(\sigma=1.5^\circ\).

We construct three evaluation settings to separately assess generalization to unseen actions and unseen material properties. \emph{ID} contains action instances that are not used for training but are drawn from the same action and material-property distributions, measuring generalization to unseen trajectories under in-distribution conditions. \emph{Action-OOD} evaluates action generalization by withholding continuous regions of the action-parameter space from training and testing exclusively on action instances from these held-out regions. For example, for fold actions, we withhold fold fractions in \([0.42,0.70]\), where the fold fraction specifies the fraction of the distance from the grasp point to the opposing edge traversed during the fold. For reposition actions, we withhold the boundary region where both the grasp and target lie in the upper-right part of their normalized coordinate spaces (\(x,y>0.69\)). For the \emph{ID} and \emph{Action-OOD} splits, assignment is performed at the action-instance level, so all material variants of the same action instance remain in the same partition. \emph{Physics-OOD} evaluates unseen material-property configurations using held-out action instances drawn from the same action distribution as training. Six material configurations are used in all training stages, while the remaining two are withheld from both tokenizer and predictive-model training and used exclusively for Physics-OOD evaluation. This preserves the action distribution while isolating the shift in material configuration. Examples of this dataset are shown in the accompanying video. The dataset available at \url{https://anonymous.4open.science/r/ClothCode-DE32}.

\subsection{Task Description}

We evaluate our framework in both simulation and real-world manipulation. In simulation, we evaluate full-state reconstruction under ID and Action-OOD settings, and action-conditioned dynamics prediction under ID, Action-OOD, and Physics-OOD settings. For dynamics, we additionally evaluate two rollout modes: \emph{Oracle}, using ground-truth context at each prediction window, and \emph{Free}, autoregressively using previous predictions as context. We further integrate the learned state and dynamics models into model-predictive control (MPC)\cite{garcia1989model} for real-world folding and flattening.

\paragraph{Folding}
Starting from an approximately flat configuration, the robot is required to transform the cloth into a prescribed folded configuration. We consider multiple folding targets, including diagonal-folds and target-specific folded shapes. 

\paragraph{Flattening.}
Starting from a crumpled configuration, the robot is required to recover an approximately flat state. Rather than following a prescribed folding geometry, the controller must select actions that progressively reduce folds and overlaps from highly variable initial configurations. 

Simulation experiments are conducted on three garment templates. A rectangular fabric, a T-shirt, and a sweater. Their dimensions are approximately \(40\times40\) cm, \(78.9\times66.2\) cm, and \(128.1\times69.2\) cm, respectively. Each template is represented by a mesh containing approximately 6,000 vertices. Real-world experiments use three fabric sheets, three T-shirts, and three sweaters with varying sizes and material properties. Within each category, all garments are mapped to the same simulated template and evaluated using the same template-specific model.

\vspace{-1mm}
\subsection{Motion Planning}
We use OWLv2~\cite{minderer2023scaling} and SAM2~\cite{ravi2025sam} to segment the garment from a single RGB-D observation and extract the corresponding partial point cloud. Our model reconstructs the complete cloth mesh, which is registered to the observation by matching its spatial extent followed by ICP refinement~\cite{besl1992method}. The target configuration is captured beforehand from an RGB-D observation and reconstructed using the same state-estimation pipeline. Following UniClothDiff~\cite{tian2025diffusion}, the grasp vertex is sampled according to the displacement between the current and target configurations, so that vertices farther from their target positions are more likely to be selected. For planning, we use CEM sampling-based MPC~\cite{de2005tutorial}. We optimize a horizon-$40$ sequence of 3-D relative end-effector displacements using 16 candidate sequences per iteration for five optimization iterations and after optimization, the robot executes the first 10 steps of the lowest-cost sequence before replanning. Each candidate sequence is rolled out with the learned dynamics model and scored by a terminal cost combining MSE and Chamfer distance to the target configuration. After execution, the garment is re-observed and its state re-estimated before the next planning cycle, forming a closed-loop perception--prediction--control loop.
\vspace{-1mm}
\section{Experiment}
% \subsection{Evaluation Metrics}
 
All errors are reported in centimeters. \emph{Mean displacement error} (MDE) measures the mean Euclidean distance between corresponding vertices. \emph{Earth Mover's Distance} (EMD) and \emph{Chamfer distance} compare meshes as point sets using optimal and nearest-neighbour matching, respectively. We report MDE and EMD for both tasks, and additionally Chamfer distance for full-state estimation to remain comparable with garment-reconstruction literature.

\subsection{Dynamics Prediction}
\begin{table}[t]
\vspace{1.5mm}
  \centering
  \caption{Action-conditioned dynamics prediction (cm, $\downarrow$). Single-window results predict 10 future frames from 3 context frames. Rollout MDE is reported on ID with oracle and predicted (\emph{free}) context.}
  \label{tab:dyn}

  % \scriptsize
  % \setlength{\tabcolsep}{7.0pt}
  % \renewcommand{\arraystretch}{0.92}
  \resizebox{\columnwidth}{!}{%
  \begin{tabular}{l cc cc cc cc}
    \toprule
    & \multicolumn{2}{c}{ID}
    & \multicolumn{2}{c}{Action-OOD}
    & \multicolumn{2}{c}{Physics-OOD}
    & \multicolumn{2}{c}{Rollout} \\
    \cmidrule(lr){2-3}
    \cmidrule(lr){4-5}
    \cmidrule(lr){6-7}
    \cmidrule(lr){8-9}
    Method
    & MDE$\downarrow$ & EMD$\downarrow$
    & MDE$\downarrow$ & EMD$\downarrow$
    & MDE$\downarrow$ & EMD$\downarrow$
    & Oracle$\downarrow$ & Free$\downarrow$ \\
    \midrule

    \multicolumn{9}{c}{\textit{Fabric}} \\
    UCD-Transformer\cite{tian2025diffusion}
    & 1.652 & 1.511 & 2.138 & 1.965 & 1.987 & 1.741 & 1.306 & 2.684 \\
    UniClothDiff
    & 1.172 & 1.118 & 1.508 & 1.465 & 1.379 & 1.328 & 1.920 & 3.916 \\
    PGND
    & 1.345 & 1.150 & 1.702 & 1.495 & 1.545 & 1.344 & 1.499 & 2.957 \\
    \textbf{ClothCode (Ours)}
    & \textbf{0.730} & \textbf{0.694}
    & \textbf{0.884} & \textbf{0.843}
    & \textbf{0.859} & \textbf{0.818}
    & \textbf{0.864} & \textbf{1.822} \\

    % \addlinespace[1pt]
    \midrule
    \multicolumn{9}{c}{\textit{T-shirt}} \\
    UCD-Transformer
    & 1.151 & 1.114 & 1.332 & 1.306 & 1.307 & 1.269 & 1.060 & 2.335 \\
    UniClothDiff
    & 1.014 & 0.984 & 1.263 & 1.247 & 1.206 & 1.175 & 1.245 & 2.675 \\
    PGND
    & 1.111 & 1.004 & 1.317 & 1.209 & 1.291 & 1.189 & 1.153 & 2.432 \\
    \textbf{ClothCode (Ours)}
    & \textbf{0.674} & \textbf{0.651}
    & \textbf{0.762} & \textbf{0.750}
    & \textbf{0.768} & \textbf{0.744}
    & \textbf{0.725} & \textbf{1.654} \\

    % \addlinespace[1pt]
    \midrule
    \multicolumn{9}{c}{\textit{Sweater}} \\
    UCD-Transformer
    & 1.452 & 1.402 & 1.659 & 1.621 & 1.794 & 1.750 & 1.096 & 2.312 \\
    UniClothDiff
    & 0.979 & 0.961 & 1.208 & 1.197 & 1.211 & 1.192 & 1.645 & 3.560 \\
    PGND
    & 0.996 & 0.890 & 1.157 & 1.145 & 1.178 & 1.070 & 1.156 & 2.337 \\
    \textbf{ClothCode (Ours)}
    & \textbf{0.653} & \textbf{0.636}
    & \textbf{0.701} & \textbf{0.694}
    & \textbf{0.806} & \textbf{0.788}
    & \textbf{0.749} & \textbf{1.362} \\
    \bottomrule
  \end{tabular}%
  }
  \vspace{-5mm}
\end{table}

\subsubsection{Baselines}
We compare ClothCode with three cloth-dynamics models. \emph{UniClothDiff}~\cite{tian2025diffusion} uses Transformer-based diffusion to predict continuous future cloth states. \emph{UCD-Transformer}~\cite{tian2025diffusion} uses the same backbone with direct MSE regression. \emph{PGND}~\cite{zhang2025particle} propagates dense particles through a spatial grid. All models use the same trajectories and splits, and predict ten future states from three context states.

\begin{figure*}[t]
    \centering
    \includegraphics[width=\textwidth]{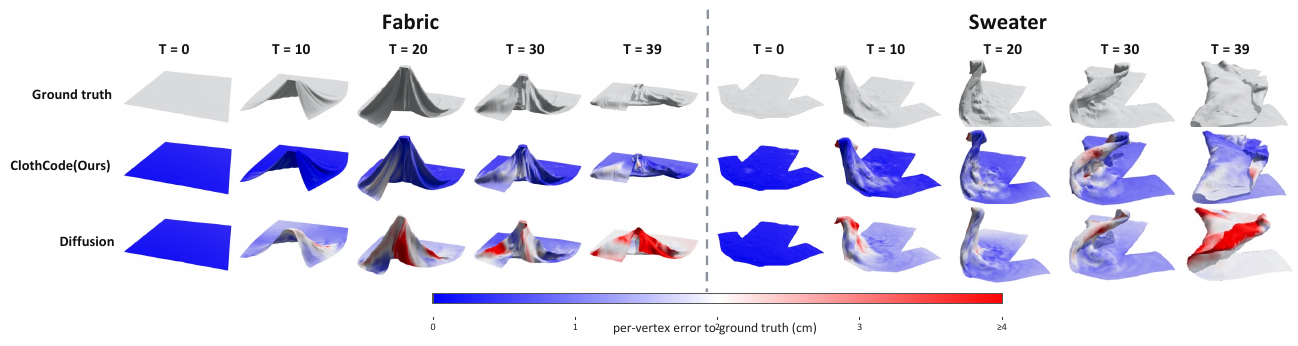}
    \caption{
    Qualitative comparison of garment dynamic prediction.
    Per-vertex reconstruction errors with respect to the ground-truth mesh are visualized using a shared color scale across all methods.
    }
    \label{fig:dynamic}
    \vspace{-2mm}
\end{figure*}

% \subsubsection{Baselines}
% We evaluate MGT against three representative cloth-dynamics models. \emph{UniClothDiff}~\cite{tian2025diffusion} uses a Transformer-based diffusion model to generate continuous future cloth states conditioned on the current state and robot actions. \emph{UCD-Transformer}\cite{tian2025diffusion} keeps the same backbone but replaces diffusion with direct MSE regression, giving a controlled deterministic ablation. We further compare with \emph{PGND}~\cite{zhang2025particle}, a particle-grid neural dynamics model that propagates dense particles through a spatial grid. All models are trained and evaluated on the same trajectories and splits, and predict ten future states from three context states.
\paragraph{Results.}
Table~\ref{tab:dyn} shows that ClothCode achieves the lowest MDE and EMD across all garments and evaluation splits, reducing them by 37.0\% and 36.2\% on average relative to the strongest baselines. The gains remain substantial under Action-OOD and Physics-OOD, with MDE reductions of 40.2\% and 35.2\%, respectively. Over multi-window rollouts, MGT reduces MDE by 32.4\% under Oracle context and 34.1\% under Free rollout. Despite using four refinement steps, ClothCode requires only 16.7\,ms per prediction window, compared with 556.2\,ms for Uniclothdiff model~\cite{tian2025diffusion} using 100 denoising steps. Fig.~\ref{fig:dynamic} further shows improved preservation of garment structure over long horizons.

\subsection{Garment state Estimation}
\label{sec:exp_pose}

\begin{figure*}[t]
    \centering
    \includegraphics[width=0.96\textwidth]{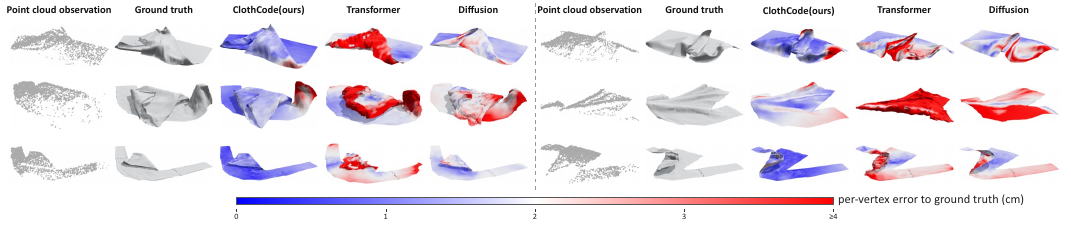}
    \caption{
    Qualitative comparison of garment state estimation from camera observations.
    Per-vertex reconstruction errors with respect to the ground-truth mesh are visualized using a shared color scale across all methods.
    }
    \label{fig:pose_qualitative}
    \vspace{-2mm}
\end{figure*}
\paragraph{Baselines.}
We compare ClothCode against the corresponding \emph{Transformer} and \emph{Diffusion} baselines described above~\cite{tian2025diffusion}, together with two garment-reconstruction methods. GarmentNets~\cite{chi2021garmentnets} reconstructs garment pose through canonical-space shape completion, while MEDOR~\cite{huang2022mesh} further refines the reconstructed mesh through test-time optimization.

\begin{table}[t]
\vspace{1.5mm}
\centering
\caption{Garment state estimation from partial point-cloud observations.}
\label{tab:pose_estimation}

\setlength{\tabcolsep}{3.2pt}
\renewcommand{\arraystretch}{0.92}

\resizebox{\columnwidth}{!}{%
\begin{tabular}{llcccccc}
\toprule
& & \multicolumn{3}{c}{Eval A (ID)}
& \multicolumn{3}{c}{Eval B (Action-OOD)} \\
\cmidrule(lr){3-5}
\cmidrule(lr){6-8}

Garment & Method
& MDE$\downarrow$
& EMD$\downarrow$
& Chamfer$\downarrow$
& MDE$\downarrow$
& EMD$\downarrow$
& Chamfer$\downarrow$ \\

\midrule

\multirow{5}{*}{Fabric}
& Uniclothdiff
& 1.384 & 1.205 & 1.770
& 1.624 & 1.352 & 1.896 \\

& Transformer
& 1.910 & 1.388 & 1.881
& 2.285 & 1.565 & 2.021 \\

& GarmentNets
& 1.993 & 1.745 & 2.314
& 2.191 & 1.870 & 2.423 \\

& MEDOR
& 2.106 & 1.809 & 2.343
& 2.290 & 1.918 & 2.417 \\

& ClothCode (Ours)
& \textbf{0.890}
& \textbf{0.680}
& \textbf{1.182}
& \textbf{1.104}
& \textbf{0.795}
& \textbf{1.279} \\

\midrule

\multirow{5}{*}{T-shirt}
& Uniclothdiff
& 1.973 & 1.690 & 2.322
& 1.956 & 1.666 & 2.301 \\

& Transformer
& 3.678 & 2.649 & 2.970
& 3.903 & 2.715 & 3.029 \\

& GarmentNets
& 2.868 & 2.495 & 3.107
& 2.976 & 2.614 & 3.259 \\

& MEDOR
& 2.998 & 2.529 & 3.025
& 3.062 & 2.609 & 3.088 \\

& ClothCode (Ours)
& \textbf{1.211}
& \textbf{0.966}
& \textbf{1.518}
& \textbf{1.173}
& \textbf{0.915}
& \textbf{1.427} \\

\midrule

\multirow{5}{*}{Sweater}
& Uniclothdiff
& 1.665 & 1.311 & 1.937
& 1.913 & 1.446 & 2.036 \\

& Transformer
& 2.637 & 1.520 & 2.070
& 3.206 & 1.649 & 2.159 \\

& GarmentNets
& 3.241 & 2.611 & 2.799
& 3.736 & 2.992 & 3.136 \\

& MEDOR
& 3.252 & 2.587 & 2.635
& 3.718 & 2.930 & 2.911 \\

& ClothCode (Ours)
& \textbf{1.452}
& \textbf{0.955}
& \textbf{1.537}
& \textbf{1.902}
& \textbf{1.139}
& \textbf{1.704} \\

\bottomrule
\end{tabular}%
}

\vspace{-2mm}
\end{table}

\paragraph{Results.}
Table~\ref{tab:pose_estimation} shows that ClothCode achieves the best performance across all garment categories, evaluation splits, and metrics, reducing MDE, EMD, and Chamfer distance by 26.6\%, 36.8\%, and 29.2\% on average relative to Uniclothdiff~\cite{tian2025diffusion}, the strongest baseline. The gains remain substantial under Action-OOD, with corresponding reductions of 23.9\%, 36.2\%, and 29.2\%. Notably, the relative improvement in EMD exceeds that in MDE for every garment under both splits, indicating stronger recovery of overall garment geometry. The improvement is particularly pronounced for the T-shirt under Action-OOD, where MDE, EMD, and Chamfer distance decrease by 40.0\%, 45.1\%, and 38.0\%, respectively. Fig.~\ref{fig:pose_qualitative} further shows that ClothCode better preserves global garment shape and folded structure under partial observation and self-occlusion.

\vspace{-2mm}
\subsection{Real-World Planning}
\label{sec:real_planning}
\begin{table}[t]
\centering
\caption{Real-world MPC success rates (\%). Each garment column
reports folding/flattening (F/F), respectively.}
\label{tab:real_mpc}
\setlength{\tabcolsep}{1.0pt}
\begin{tabular}{lccc}
\toprule
State + Dynamics & Fabric F/U & T-shirt F/U & Sweater F/U \\
\midrule
ClothCode + Uniclothdiff
    & 83.3/76.7
    & 66.7/70.0
    & 50.0/56.7 \\
Uniclothdiff + ClothCode
    & 73.3/73.3
    & 70.0/70.0
    & 60.0/66.7 \\
Uniclothdiff + Uniclothdiff
    & 66.7/66.7
    & 56.7/60.0
    & 40.0/50.0 \\
ClothCode + ClothCode (Ours)
    & \textbf{90.0/80.0}
    & \textbf{80.0/80.0}
    & \textbf{66.7/76.7} \\
\bottomrule
\end{tabular}
\vspace{-5mm}
\end{table}

\begin{figure*}[t]
    \centering
    \includegraphics[width=0.85\textwidth]{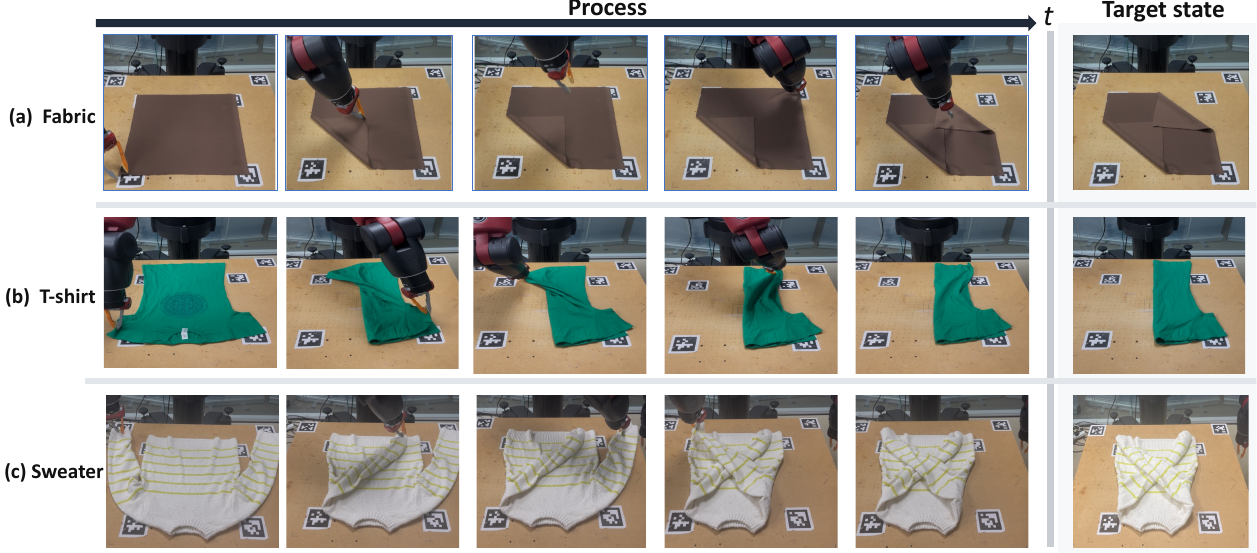}
    \caption{Qualitative results of real-world experiments.}
    \label{fig:real}
    \vspace{-5mm}
\end{figure*}

% Using a Baxter robot and a single ZED 2i camera, we evaluate all four combinations of MGT- and diffusion-based pose estimation and dynamics prediction while keeping the MPC objective, and execution fixed. We test three garments from each garment category on folding and unfolding, with ten trials per instance. For each garment instance, we use the final goal state as the target. A trial is considered a \emph{success} if the final-mask IoU with the target is at least $0.80$.

% As shown in Table~\ref{tab:real_mpc}, combining MGT pose estimation with MGT dynamics achieves the highest success rate in all six garment--task settings: \(90.0/80.0\%\) on fabric, \(80.0/80.0\%\) on T-shirts, and \(66.7/76.7\%\) on sweaters for folding/unfolding, respectively, corresponding to \(78.9\%\) overall. Replacing only the pose estimator or dynamics model with its diffusion counterpart reduces the overall success rate to \(68.9\%\) and \(67.2\%\), respectively, while replacing both reduces it to \(56.7\%\). The mixed variants reveal a category-dependent bottleneck. With MGT dynamics fixed, MGT pose estimation improves the average fabric success rate from \(73.3\%\) to \(85.0\%\), whereas with MGT pose estimation fixed, MGT dynamics improves sweater performance from \(53.3\%\) to \(71.7\%\). This suggests that state estimation is the larger bottleneck for geometrically simpler fabric, while accurate multi-step dynamics becomes increasingly important for complex, multi-layer garments.

Using a Baxter robot and a single ZED 2i camera, we evaluate all four combinations of ClothCode and Uniclothdiff~\cite{tian2025diffusion} state estimation and dynamics prediction under the same MPC and execution settings. We test three garments per category on folding and flattening, with ten trials per instance, and define \emph{success} as a final-mask IoU of at least $0.80$ with the target. As shown in Table~\ref{tab:real_mpc}, ClothCode state estimation and dynamics prediction achieves the highest success rate in all six garment--task settings, reaching $78.9\%$ overall. Replacing the state estimator or dynamics model with Uniclothdiff reduces overall success to $68.9\%$ and $67.2\%$, respectively, while replacing both reduces it to $56.7\%$. The mixed variants also reveal a category-dependent bottleneck: with ClothCode dynamics fixed, ClothCode state estimation improves fabric success from $73.3\%$ to $85.0\%$, whereas with ClothCode state estimation fixed, ClothCode dynamics improves sweater success from $53.3\%$ to $71.7\%$. Some qualitative results of real-world experiments are shown in Fig.~\ref{fig:real}.

\vspace{-1mm}
\subsection{Ablation}
% \paragraph{Effect of residual quantization.}
% We ablate the discrete representation by comparing a continuous patch latent, single-level vector quantization (VQ), and residual vector quantization (RVQ) under the same patch-wise representation. The continuous tokenizer provides an upper bound on representation fidelity, achieving a validation MDE of $0.175$ cm. Replacing the continuous latent with a single $512$-entry VQ codebook increases the error to $0.367$ cm, indicating a substantial quantization bottleneck when each patch is represented by only one discrete code. Using two residual quantization levels reduces the error to $0.187$ cm
% and the P95 error from $1.503$ cm to $0.787$ cm, recovering most of the continuous representation accuracy. These results show that residual quantization substantially improves the fidelity of the discrete patch representation by allowing multiple code vectors to jointly approximate each patch latent, without increasing the number of material patches.

\paragraph{Effect of residual quantization.}
Under the same patch-wise representation, continuous latents achieve $0.175$ cm MDE, while single-level VQ increases the error to $0.367$ cm. Two-level RVQ reduces MDE to $0.187$ cm and P95 error from $1.503$ cm to $0.787$ cm, recovering most of the continuous representation fidelity.

\paragraph{Representation and prediction design.}
We ablate patch-wise factorization, discretization, and iterative generation as shown in Table~\ref{tab:ablation}. The discrete one-shot baseline is trained independently for single-pass token prediction rather than obtained by truncating the masked model at inference. As shown in Table~\ref{tab:ablation}, discretization reduces state MDE from $2.673$ to $1.996$~cm under one-shot prediction while remaining competitive on dynamics ($1.327$ vs.~$1.300$~cm). Masked iterative prediction further reduces MDE to $0.674$~cm for dynamics and $1.211$~cm for state, improving over discrete one-shot by $48.2\%$ and $39.3\%$, respectively. Compared with continuous iterative prediction, the discrete masked model further reduces MDE by $33.5\%$ and $42.5\%$. These results show that discretization and masked refinement provide complementary gains.
% \paragraph{}
% \label{sec:ablation}

% We ablate the representation and prediction design varying three components: patch-wise factorization,
% discretization, and iterative generation. The discrete one-shot baseline is trained independently for single-pass token prediction and is not obtained by truncating the masked iterative model at inference. Table~\ref{tab:ablation}
% reports MDE for both dynamics prediction and garment state estimation. The discrete representation provides a structured prediction space that benefits both prediction and refinement. Under one-shot prediction, discretization reduces pose MDE from $2.673$ to $1.996$~cm while remaining competitive on dynamics ($1.327$ vs.~$1.300$~cm). Combined with masked iterative prediction, MDE further decreases to $0.674$~cm for dynamics and $1.211$~cm for pose, corresponding to reductions of $48.2\%$ and $39.3\%$ over discrete one-shot prediction. Continuous iterative diffusion also improves over continuous one-shot prediction, but the discrete masked model remains substantially more accurate, reducing MDE by $33.5\%$ on dynamics and $42.5\%$ on pose relative to continuous iterative prediction. These results show that discrete representation and masked refinement are complementary: the former provides a compact structured space for garment configurations, while the latter exploits this structure through confidence-guided iterative correction.

\begin{table}[t]
\vspace{1.5mm}
\centering
\caption{
Ablation of representation and prediction design.
MDE is reported in cm; lower is better.
}
\label{tab:ablation}
\resizebox{\columnwidth}{!}{
\begin{tabular}{llcc}
\toprule
\textbf{Representation} & \textbf{Prediction} &
\textbf{Dynamics $\downarrow$} & \textbf{State $\downarrow$} \\
\midrule
Continuous global
& One-shot regression
& $1.685 \pm 0.028$
& $2.093 \pm 0.071$ \\

Continuous patch latent
& One-shot regression
& $1.327 \pm 0.015$
& $2.673 \pm 0.070$ \\

Continuous patch latent
& Iterative diffusion
& $1.014 \pm 0.032$
& $2.105 \pm 0.170$ \\

Discrete RVQ
& Trained one-shot
& $1.300 \pm 0.019$
& $1.996 \pm 0.073$ \\

\textbf{Discrete RVQ}
& \textbf{Masked iterative}
& $\mathbf{0.674 \pm 0.023}$
& $\mathbf{1.211 \pm 0.075}$ \\
\bottomrule
\end{tabular}
}
\vspace{-7mm}
\end{table}

%\vspace{-8mm}
\section{Conclusion}
We presented \textbf{ClothCode}, a unified generative framework for garment state estimation and action-conditioned dynamics prediction in a shared discrete configuration space. ClothCode represents complete garment geometry using material-aligned RVQ tokens and predicts these tokens with task-specific Masked Generative Transformers through parallel, confidence-guided refinement. Across three garment templates and both in-distribution and out-of-distribution evaluation, ClothCode reduces mean displacement error by 26.6\% for state estimation and 37.0\% for dynamics prediction relative to the strongest baselines, while dynamics inference requires only 16.7\,ms per prediction window compared with 556.2\,ms for diffusion. ClothCode increase real-world folding and flattening success from 56.7\% to 78.9\%. %Ablations further show that the discrete bottleneck retains garment geometry with sub-centimetre reconstruction error and that iterative masked refinement improves prediction within the learned discrete space.
Future work will investigate transferable discrete garment representations across unseen garment topologies and their use in longer-horizon closed-loop manipulation.

\bibliographystyle{IEEEtran}
\bibliography{references}

\end{document}